\pdfoutput=1

\documentclass[11pt]{article}

\usepackage{EMNLP2022}

\usepackage{times}
\usepackage{latexsym}
\usepackage{graphicx}
\usepackage{pythonhighlight}
\usepackage{subfigure}
\usepackage[inkscapelatex=false]{svg}
\usepackage{multicol}
\usepackage{amssymb}
\usepackage{graphicx} 
\usepackage{natbib}  
\usepackage{caption} 
\usepackage{amssymb}
\usepackage{color}
\usepackage{algpseudocode}
\usepackage[ruled,linesnumbered]{algorithm2e}

\usepackage{pgfplots}
\usetikzlibrary{pgfplots.groupplots}
\pgfplotsset{compat=1.3}
\usepackage{tikz}
\usetikzlibrary{patterns}

\usepackage{amsmath}
\usepackage{amsthm}
\usepackage{makecell}
\usepackage{color}
\usepackage{adjustbox}
\usepackage{multirow}
\usepackage{booktabs}
\usepackage{amsmath}
\usepackage{hyperref}
\usepackage[inkscapelatex=false]{svg}
\usepackage{subfloat}

\usepackage{dsfont}
\usepackage{arydshln}
\usepackage{enumitem}

\graphicspath{{./images/}} 
\usepackage{latexsym}
\usepackage[T1]{fontenc}

\usepackage[utf8]{inputenc}

\usepackage{microtype}

\usepackage{inconsolata}

\title{Revisiting Grammatical Error Correction Evaluation and Beyond}

\author{Peiyuan Gong$^1$~~~
        Xuebo Liu$^2$\thanks{~~Corresponding author}~~~
        Heyan Huang$^1$~~~
        Min Zhang$^2$\\
  $^1$School of Computer Science and Technology, Beijing Institute of Technology, Beijing, China \\
      \texttt{\{3120201019,hhy63\}@bit.edu.cn} \\
    $^2$Institute of Computing and Intelligence, Harbin Institute of Technology, Shenzhen, China \\
          \texttt{\{liuxuebo,zhangmin2021\}@hit.edu.cn} }
          
\begin{document}
\maketitle
\begin{abstract}
Pretraining-based (PT-based) automatic evaluation metrics (e.g., BERTScore and BARTScore) have been widely used in several sentence generation tasks (e.g., machine translation and text summarization) due to their better correlation with human judgments over traditional overlap-based methods. Although PT-based methods have become the de facto standard for training grammatical error correction (GEC) systems, GEC evaluation still does not benefit from pretrained knowledge. 
This paper takes the first step towards understanding and improving GEC evaluation with pretraining. 
We first find that arbitrarily applying PT-based metrics to GEC evaluation brings unsatisfactory correlation results because of the excessive attention to inessential systems outputs (e.g., unchanged parts). 
To alleviate the limitation, we propose a novel GEC evaluation metric to achieve the best of both worlds, namely PT-M$^{2}$, which only uses PT-based metrics to score those corrected parts. 
Experimental results on the CoNLL14 evaluation task show that PT-M$^{2}$ significantly outperforms existing methods, achieving a new state-of-the-art result of 0.949 Pearson correlation. Further analysis reveals that PT-M$^{2}$ is robust to evaluate competitive GEC systems. Source code and scripts are freely available at \href{https://github.com/pygongnlp/PT-M2}{https://github.com/pygongnlp/PT-M2}.
\end{abstract}

\section{Introduction}
Grammatical error correction (GEC) is {the} task that takes a sentence with grammatical errors as input, and outputs a corrected sentence.
Due to the important role of GEC in the field of second language learning and intelligent writing, GEC has attracted wide attention from the community~\citep{chollampatt2018multilayer, lewis2019bart, omelianchuk2020gector}.
As a typical natural language generation task (NLG), the common practice for GEC evaluation is to calculate the similarity between system outputs and their corresponding references~\citep{dahlmeier2012better, napoles2015ground, bryant2017automatic}.

\begin{table}[t]
\resizebox{\linewidth}{!}{
\begin{tabular}{lcc}
\toprule
\textbf{Task} & \textbf{Training} & \textbf{Evaluation} \\ 
\midrule
Machine Translation     & $\bullet$        & $\bullet$       \\
Text Summarization     & $\bullet$        & $\bullet$       \\
Grammatical Error Correction     & $\bullet$        & $\circ$       \\ 
\bottomrule
\end{tabular}}
\caption{The use of PT-based models for model training and evaluation in various NLG tasks. $\bullet$ means used and $\circ$ means unused. Unlike other tasks, GEC has not used PT-based methods for evaluation.}
\label{pt_usage}
\end{table}

With the rapid development of pretraining (PT) \citep{devlin2018bert, liu2019roberta, lewis2019bart}, several NLG tasks, such as machine translation, text summarization and GEC, have been utilizing PT-based models to improve the model training process~\citep{kaneko2020encoder, omelianchuk2020gector, tarnavskyi2022ensembling}, as shown in Table \ref{pt_usage}.
Furthermore, since PT-based models can learn rich syntactic and semantic knowledge from a large amount of unlabeled data, they are able to calculate the similarity between two sentences more accurately.
Therefore, many mainstream NLG tasks try to develop new evaluation metrics based on PT-based models, and the new metrics have been sufficiently validated in terms of the high consistency with human judgments~\citep{zhang2019bertscore, yuan2021bartscore}.
However, existing GEC systems still use traditional metrics for evaluation. 

In this paper, to find the reason why GEC systems do not use PT-based metrics for evaluation, we revisit existing GEC evaluation, comparing the traditional overlap-based evaluation metrics with recently popular PT-based metrics.
Surprisingly, our preliminary experiment on the CoNLL14 evaluation task shows that arbitrarily applying PT-based metrics to GEC evaluation results in a relatively worse correlation to human judgments than the traditional metrics.
Further analysis reveals that the scoring strategies of PT-based metrics are questionable for GEC evaluation.
GEC is a local substitution task that only partially changes from the source sentences, but PT-based metrics have to compute the score of the whole sentence despite most words staying the same after correction.
The scores from the unchanged words bias the final sentence score, leading to an unreliable evaluation for GEC.

To alleviate the above limitation, in this paper we propose PT-M$^{2}$, a novel PT-based GEC metric that combines the advantages of both the PT-based metrics and traditional metrics.
Unlike the PT-based metrics scoring a whole sentence, PT-M$^{2}$ only uses PT-based metrics to score the changed words that can be extracted by the M$^{2}$ metric~\citep{dahlmeier2012better}.
Experimental results on the CoNLL14 evaluation task show that PT-M$^{2}$ achieves the highest correlation compared to traditional metrics based on two kinds of system ranking methods, either calculating score at the corpus- or sentence-level.
We also show that PT-M$^{2}$ does not heavily rely on the scale of PT-based models, even equipped with a small-scale model can obtain satisfactory results.

Our main contributions are listed as follows:
\begin{itemize}
    \item We revisit GEC evaluation and find that the arbitrary use of PT-based metrics results in poor correlation with human judgments. We analyze the corresponding reasons in terms of scoring strategies for different metrics.
    \item We propose a novel PT-based GEC metric PT-M$^{2}$, which only uses PT-based metrics to score the correction words. PT-M$^{2}$ achieves a new state-of-the-art of 0.949 Pearson correlation on the CoNLL14 evaluation task.
    \item We find that PT-M$^{2}$ still performs well in evaluating high-performing GEC systems, which is helpful for promoting GEC research and upgrading the GEC system in the future.
\end{itemize}
\begin{table}[t]
\centering
\resizebox{\linewidth}{!}{
\begin{tabular}{lccc}
\toprule
\textbf{Latest GEC Work} &\bf M$^{2}$ &\bf ERRANT &\bf GLEU\\
\midrule
\citet{omelianchuk2020gector} & $\bullet$   & $\bullet$     & $\circ$   \\
\citet{sun2022adjusting}  & $\bullet$   & $\bullet$     & $\circ$   \\
\citet{lai2022type} & $\bullet$   & $\bullet$     & $\circ$   \\
\citet{tarnavskyi2022ensembling}& $\bullet$   & $\bullet$     & $\circ$   \\
\hdashline
\citet{stahlberg2020seq2edits}  & $\bullet$   & $\bullet$ & $\bullet$ \\
\citet{kaneko2020encoder}   & $\bullet$   & $\bullet$ & $\bullet$ \\
\citet{katsumata2020stronger}   & $\bullet$   & $\bullet$ & $\bullet$ \\
\citet{parnow2021grammatical}  & $\bullet$   & $\bullet$ & $\bullet$ \\
\bottomrule
\end{tabular}}
\caption{\label{recent_work}
GEC metrics {which are} used in {the} latest GEC works.
All of the works evaluate with traditional GEC metrics, despite using PT-based models for training.
}
\end{table}

\section{Related Work}
\label{sec:related work}

\subsection{Overlap-based GEC Metrics}
As a growing number of high-performance GEC models are proposed~\citep{liu2021understanding,li-etal-2022-ode,Zhang2022BLISSRS}, it is important to provide interpretable and reliable evaluation metrics to measure their quality. 
Several overlap-based GEC metrics have been provided to evaluate GEC models.
As shown in Table \ref{recent_work}, despite employing PT-based models for model training, recently published works still use overlap-based GEC metrics (e.g., M$^{2}$ \cite{dahlmeier2012better}, GLEU \cite{napoles2015ground}, and ERRANT \cite{bryant2017automatic}) for GEC evaluation.

\citet{dale2011helping} first align source sentences and hypothesis sentences using the Levenshtein algorithm for GEC evaluation.
\citet{dahlmeier2012better} convert each source-hypothesis pair to an edit sequence dynamically, extracting the corresponding edits and using the F$_1$ measure to represent the score of each system. Similar to \cite{dahlmeier2012better}, \citet{felice2015towards} evaluate corrections at the token level, leveraging a globally optimal alignment algorithm {on} source-hypothesis pairs and source-reference pairs {respectively}.
\citet{napoles2015ground} propose a variant of BLEU that rewards n-grams appearing in hypothesis sentences and reference sentences but not in source sentences, penalizing n-grams in source sentences and hypothesis sentences but not in reference sentences and averaging scores over different references.
\citet{bryant2017automatic} {employ} a linguistically-enhanced Damerau-Levenshtein algorithm to align sentence pairs, {merging parts of the alignment and }automatically classifying each edit with pre-defined {rules}.
\citet{choshen2020classifying} propose a multilingual variant of ERRANT, extracting and classifying edits with universal dependencies. 
\citet{gotou2020taking} focus on difficulty for the model to correct different types of errors and use several GEC models to compute the difficulty weight for each edit.
However, a natural weakness is that traditional overlap-based metrics cannot capture the similarity between semantically similar words, limiting their effectiveness.

\subsection{PT-based NLG Metrics}
PT-based methods are not only used in the training stage of NLG tasks but also in designing evaluation metrics to assess the performance of NLG models, which can overcome the above core weakness of overlap-based metrics.
Several recently proposed PT-based metrics have dominated a large number of NLG evaluation tasks. 
\citet{lo2019yisi} use cross-lingual PT-based models to encode source sentences and hypothesis sentences and compute the cosine distances with a semantic parser. 
\citet{zhao2020limitations} compute scores for source-hypothesis pairs with cross-lingual PT-based models, re-aligning the vector space and using a language model to score fluency. 
\citet{belouadi2022uscore} develop a fully unsupervised evaluation metric that leverages pseudo-parallel pairs obtained from fully unsupervised evaluation metrics and {use} pseudo reference sentences from unsupervised translation systems. 
\citet{song2021sentsim} introduce a metric combined by BERTScore, {Word Mover's Distance} and sentence semantic similarity. 
\citet{gekhman2020kobe} extract the entities from source sentences and hypothesis sentences via a multilingual knowledge base respectively and measure the recall of two entitie sets.
\citet{yoshimura2020some} train a BERT-based regression model to optimize multiple sub-metrics on the GEC evaluation dataset.
\citet{islam2021end} leverage GPT2 \cite{radford2018improving} to measure the grammaticality for the GEC outputs.

\citet{zhang2019bertscore} compute the similarity score for each token in hypothesis sentences with each token in reference sentences and use greedy matching to maximize each similarity score. \citet{zhao2019moverscore} combine contextualized embeddings with Word Mover's Distance to soft align tokens from hypothesis sentences to reference sentences. \citet{yuan2021bartscore} introduce that a high-quality hypothesis will be easily generated based on source sentence or reference sentence or vice-versa, and use the generation probability to evaluate system outputs. 
\citet{sellam2020bleurt} generate a large number of synthetic hypothesis-reference sentence pairs and pretrain BERT on several supervision signals with a parameterized regression layer to help {the} model generalization. 
\citet{rei2020comet} propose two evaluation frameworks that predict the quality score and minimize the distance between the ``better'' hypothesis sentence and the corresponding reference sentence respectively. \citet{zhan2021varianceaware} evaluate different PT-based NLG metrics for machine translation tasks.
Benefiting from PT-based metrics, the correlation with human judgments has significantly improved on NLG evaluation. In this work, we would like to investigate the effectiveness of PT-based metrics on GEC evaluation.

\begin{table*}[t]
\small
{
\begin{tabular}{llcc}
\toprule
\textbf{System} &\textbf{Sentence} & \textbf{Rank} & \textbf{BERTScore} \\ 
\midrule
\textbf{SRC} &  They play \textcolor{red}{the} important role in our life which can not be \textcolor{blue}{\bf substituted} .
&- & -      \\
\textbf{REF} & They play \textcolor{red}{\bf an}  important role in our life which can not be \textcolor{blue}{\bf substituted} .
&- & - \\
\midrule
\textbf{AMU} & They play \textcolor{red}{\bf an (0.99)} important role in our life which can not be \textcolor{blue}{replaced (0.75)} .
& 1 & 0.94 \\
\textbf{UFC} & They play \textcolor{red}{the (0.63)} important role in our life which can not be 
\textcolor{blue}{\bf substituted (0.99)} . 
& 2 & {\textbf{0.95}} \\ 

\bottomrule
\end{tabular}}
\caption{Example from the CoNLL14 evaluation task. ``\textcolor{red}{\bf Red Bold}'' denotes the right correction whereas ``\textcolor{blue}{Blue Non-bolded}'' denotes the wrong one. Although BERTScore scores higher for the correct edit of the AMU system, the final overall score is worse than that of the UFC system.}
\label{pt_example}
\end{table*}

\section{Revisiting Existing Metrics}
This section revisits existing overlap-based GEC metrics and PT-based metrics on the GEC evaluation task, computing the metric correlation with human judgments, analyzing experimental results and exploring the differences among the metrics.
\label{sec:revisit}

\subsection{Experimental Setup}
\label{dataset}
\paragraph{Dataset Settings}
We conduct experiments on the CoNLL14 evaluation task \cite{grundkiewicz2015human}.
There are 1,312 source sentences, and each source sentence corresponds to two standard reference sentences. 
Twelve teams have provided their system outputs and the source sentences are used as the thirteenth system. 
Eight annotators judge the quality of hypothesis sentences generated by corresponding GEC systems. Each hypothesis sentence is scored from 1 to 5, which means from worst to best. 
Two system ranking lists are respectively generated by Expected
-Wins (EW) and Trueskill (TS) algorithms.

\paragraph{Experiment Settings}
\label{evaluation_settings}
The method to estimate the performance of GEC evaluation metrics is measuring the degree of correlation with human judgments. 
Following \cite{grundkiewicz2015human}, we measure the correlation between metrics and human judgments based on the \textbf{system-level ranking}, computing Pearson $\gamma$ and Spearman $\rho$ respectively as the final correlations. 
We compute the score for each system in two settings: \textbf{corpus-level} and \textbf{sentence-level}. 
Given the metric $\text{M}$, source sentences $\mathbf{S}$, hypothesis sentences $\mathbf{H}$ and reference sentences $\mathbf{R}$, the first setting computes the system score based on the whole corpus $\text{M}(\mathbf{S}, \mathbf{H}, \mathbf{R})$, and the last one uses the average of the sentence-level scores $\sum_{i}^{I}\text{M}(\mathbf{S}_i, \mathbf{H}_i, \mathbf{R}_i)/ I$.

\paragraph{Evaluation Metrics}
We compare the following overlap-based GEC metrics and PT-based metrics:
\label{metrics}
\begin{itemize}
\item \textbf{GLEU} rewards hypothesis n-grams that match reference sentences but not source sentences and penalizes hypothesis n-grams that match source sentences but not reference sentences \cite{napoles2015ground}. 
\item \textbf{M$^{2}$} aligns source sentences and hypothesis sentences with Levenshtein algorithm, dynamically choosing the alignment that maximally matches the gold edits and extracting the system edits. \cite{dahlmeier2012better}. F$_{0.5}$ is used as the system score.
\item \textbf{SentM$^{2}$}  is a variant of M$^{2}$, using the average of F$_{0.5}$ scores computed at the sentence-level as the system score.
\item \textbf{ERRANT} aligns sentence pairs with a linguistic-enhanced Damerau-Levenshtein algorithm and uses two kinds of rules to merge alignment, extract and classify edits \cite{bryant2017automatic}. F$_{0.5}$ is used as the system score.
\item \textbf{SentERRANT} is a variant of ERRANT, using the average of F$_{0.5}$ scores computed at the sentence-level as the system score.
\item \textbf{BERTScore} is a PT-based metric, which computes the token similarity between the sentence pairs and uses greedy matching to maximize the matching similarity score \cite{zhang2019bertscore}.\footnote{We use the reference sentences from the CoNLL14 evaluation task as the corpus to compute the IDF weight.}
\item \textbf{BARTScore} is a PT-based metric, which converts the evaluation task to a sequence generation task and uses the generation probability to estimate the quality of system outputs \cite{yuan2021bartscore}.\footnote{We use the reference sentence as the input and generate the corresponding hypothesis sentence.}
\end{itemize}

\begin{figure*}[t]
\centering
\includegraphics[width=\linewidth]{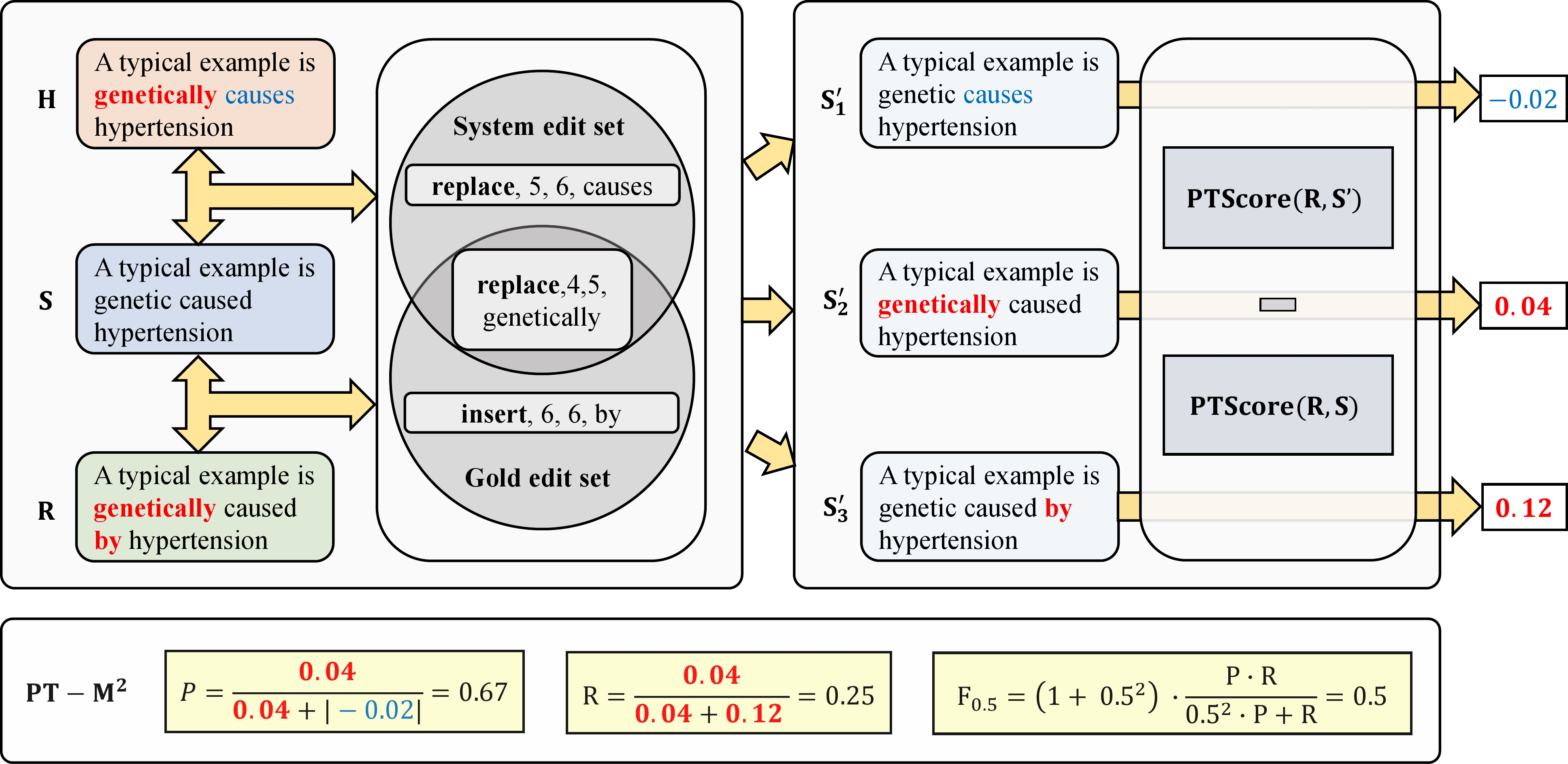}
\caption{Overview of our approach \textbf{PT-M$^{2}$}, $\mathbf{S}$, $\mathbf{H}$, and $\mathbf{R}$ denote the source sentence, the hypothesis sentence, and the reference sentence, respectively. There are three core modules in our approach: 1) Extract two edit sets from corresponding sentence pairs; 2) Compute the score for each edit in the edit set with PT-based metrics; 3) Apply edit scores as corresponding edit weights on the overlap-based GEC metrics M$^{2}$. ``\textcolor{red}{\bf Red Bold}'' denotes the right corrections and their corresponding scores whereas ``\textcolor{blue}{Blue Non-bolded}'' denotes the wrong corrections and their corresponding scores.}
\label{fig1}
\end{figure*}

\begin{table}[t]
\centering
\scalebox{0.95}{
\begin{tabular}{@{}lcccc@{}}
\toprule
\multirow{2}{*}{\textbf{Metric}} & \multicolumn{2}{c}{\textbf{EW}} & \multicolumn{2}{c}{\textbf{TS}} \\ 
\cmidrule(lr){2-3} \cmidrule(lr){4-5}
& $\gamma$ & $\rho$ & $\gamma$ & $\rho$   \\
\midrule   
GLEU & 0.701 & 0.467 & 0.750 & 0.555 \\
ERRANT & 0.642 & 0.659 & 0.688 & 0.698 \\
SentERRANT & 0.870 & \textbf{0.742} & 0.846 & 0.747 \\
M$^{2}$ & 0.623 & 0.687 & 0.672 & 0.720 \\
SentM$^{2}$ & \textbf{0.871} & 0.731 & \textbf{0.864} & \textbf{0.758} \\
\midrule    
BARTScore & 0.172 & 0.253 & 0.173 & 0.269 \\
BERTScore & 0.262 & 0.074 & 0.166 & -0.022 \\
\bottomrule       
\end{tabular}}        
\caption{Correlations of metrics with human judgments on the CoNLL14 evaluation task. PT-based metrics such as BERTScore and BARTScore perform relatively worse than the overlap-based GEC metrics.}
\label{exist_result}    
\end{table}  

\subsection{Preliminary Results}  
\label{baseline_experiment}
\paragraph{PT-based Metrics Fail} The results of the existing metrics are shown in Table \ref{exist_result}. The advantages are different between GLEU and M$^{2}$ as GLEU is better than M$^{2}$ on Pearson and worse on Spearman \cite{chollampatt2018reassessment}. We introduce ERRANT to compute its correlation with human judgments since ERRANT can extract and classify edits automatically, which has been used as the standard metric in GEC \cite{bryant2017automatic}. Without using the gold edits annotated by humans, ERRANT can still get a relatively high correlation. We also find that computing the system score at the sentence-level for M$^{2}$ and ERRANT can get a higher correlation than computing at the corpus-level, which is the same as in \cite{napoles2016there}.

Besides, we observe that even though the PT-based metrics such as BERTScore and BARTScore have dominated in the evaluation of multiple NLG tasks \cite{scialom2021beametrics}, they correlate much lower than overlap-based GEC metrics on the GEC evaluation task, whether for Pearson or Spearman, and are even negatively correlated with human judgments. So why are PT-based metrics not suitable to evaluate GEC systems? What are the differences between the PT-based metrics and overlap-based GEC metrics? 

\paragraph{Discussion}      
\label{discussion}    
This part aims to answer the above questions. As shown in Table~\ref{pt_example}, to figure out why PT-based metrics are not suitable for GEC evaluation, we analyze a representative from the CoNLL14 evaluation task. 
For the two edits provided by the AMU system, even though the change of the correct edit score is larger than that of the wrong correction edit, the AMU system score provided by BERTScore is lower than that of the UFC system, which leads to the result being different with human annotation. 
A possible reason is that BERTScore computes scores for all the tokens in the sentence, a large percentage of the score for the unchanged part affects the trend of the overall score. 
When an edit is applied, not only the corresponding edit score is changed, but the scores of surrounding tokens are also affected, causing BERTScore to fail to make a correct assessment.
The excessive attention to unchanged words potentially biases the final score of PT-based metrics.

\section{PT-M$^2$: The Best of Both Worlds}
\label{sec:method}  
\subsection{Motivation}
\label{motivation}

Based on the above findings, to inject pretrained knowledge into GEC evaluation metrics, we propose PT-M$^{2}$, which takes advantage of both PT-based metrics and overlap-based GEC metrics.
PT-M$^{2}$ computes edit scores with PT-based metrics. 
Without directly using PT-based metrics to score hypothesis-reference sentence pairs, we use them at the edit-level to compute a score for each edit. 
As shown in Figure \ref{fig1}, we first extract the system edit set and gold edit set from the source-hypothesis sentence pair and source-reference sentence pair respectively. We then compute the score of each edit with PT-based metrics as the scorer. Lastly we apply edit scores as the edit weights on the M$^{2}$.

\subsection{Implementation}
\paragraph{Edit Extraction}
\label{edit_gec_metrics}
Given a source sentence $S$, a hypothesis sentence $H$ and a reference sentence $R$, the first step we have to do is to extract the system edit set $E$ and gold edit set $G$ from the source-hypothesis sentence pair $(S, H)$ and source-reference sentence pair $(S, R)$ respectively \cite{dahlmeier2012better, bryant2017automatic}. 
As shown in Figure \ref{fig1}, each edit is composed of edit operation, start index, end index and correct tokens.
We use the edit extraction module from M$^{2}$ to extract both two edit sets.\footnote{Due to the fact that the CoNLL14 evaluation task has provided the gold edit set, we only need to extract the system edit set by M$^{2}$.} The intersection of the system edit set and gold edit set represents all of the correct edits provided by corresponding system.

\begin{table*}[t]
\scalebox{0.93}{
\centering
{
\begin{tabular}{llccccccccr}
\toprule
\multirow{2}{*}{\textbf{Type}} & \multirow{2}{*}{\textbf{Metric}} & \multirow{2}{*}{\textbf{PT Model}}& 
\multicolumn{2}{c}{\textbf{EW (Corpus)}} & \multicolumn{2}{c}{\textbf{TS (Corpus)}} &
\multicolumn{2}{c}{\textbf{EW (Sentence)}} & \multicolumn{2}{c}{\textbf{TS (Sentence)}} \\
\cmidrule(lr){4-5} \cmidrule(lr){6-7} \cmidrule(lr){8-9} \cmidrule(lr){10-11} 
& & &$\gamma$ & $\rho$ & $\gamma$ & $\rho$ & $\gamma$ & $\rho$ & $\gamma$ &\multicolumn{1}{c}{$\rho$} \\

\midrule 
\multirow{3}{*}{Overlap} & GLEU & - & 0.701 & 0.467 & 0.750 & 0.555 & 0.784 & 0.720 & 0.828 & 0.775\ \\ 
& ERRANT & - & 0.642 & 0.659 & 0.688 & 0.698 & 0.870 & 0.742 & 0.846 & 0.747 \\ 
& M$^{2}$ & - & 0.623 & 0.687 & 0.672 & 0.720 & 0.871 & 0.731 & 0.864 & 0.758 \\ 

\midrule 
\multirow{2}{*}{PT} & BARTScore & BART & - & - & - & - & 0.172 & 0.253 & 0.173 & 0.269 \\ 
& BERTScore & BERT & - & - & - & - & 0.262 & 0.074 & 0.166 &-0.022 \\ 
\midrule 
\multirow{4}{*}{Ours}& PT-ERRANT & BART & 0.681 &\bf 0.797 & 0.727 &\bf 0.841 & 0.905 & 0.786 & 0.897 & 0.824 \\ 
& PT-ERRANT & BERT &\bf 0.705 & \underline{0.780} &\bf 0.745 & 0.797 & \underline{0.941} & \underline{0.879} & \underline{0.917} & \underline{0.846} \\
\cdashline{2-11}
& PT-M$^{2}$ & BART & 0.667 & 0.775 & 0.715 & \underline{0.813} & 0.898 & 0.813 & 0.901 & \underline{0.841} \\ 
& PT-M$^{2}$ & BERT & \underline{0.693} & 0.758 &\underline{0.737} & 0.769 & \textbf{0.949} & \textbf{0.907} & \textbf{0.938} & \bf{0.874} \\ 
\bottomrule
\end{tabular}}}
\caption{Correlations of metrics with human judgments on the CoNLL14 evaluation task. Our proposed PT-M$^{2}$ and PT-ERRANT methods correlate better with human judgments on both the corpus- and sentence-level. We highlight the \textbf{highest} score in bold and the \underline{second-highest} score with underlines. }
\label{main_result}
\end{table*}

\paragraph{Edit Score}
\label{editscore}
Without treating each edit equally, we propose a novel method to score each edit, employing PT-based metrics (e.g., BERTScore and BARTScore) as the edit scorer (PTScore). 
As shown in Figure~\ref{fig1}, we first build an edit set $U$, which is the union of the system edit set $E$ and gold edit set $G$. 
Then, we compute the score of each edit $u$ in the edit set $U$ and obtain an edit score set $W$, in which $w_u$ is the score of the edit $u$.
To achieve this goal, the first step we applied is to use each edit {in the edit set $U$} to generate a partially correct version {$S'$} of the source sentence $S$. 
Then we use PTScore to compute the scores of the sentence pair $(S, R)$ and {$(S', R)$} respectively, the more similarity between the sentence pair which input to PTScore, the higher the score that PTScore gives. 
The score differences are used to measure whether the edit is beneficial or not:
\begin{equation}
    w = \text{PTScore}(S', R) - \text{PTScore}(S, R)
\end{equation}
$w$ > 0 means the edit is helpful for correction, which shows it is a correct edit, otherwise is a wrong correction edit. We use the absolute value $\left|w\right|$ as the edit score. The larger the $\left|w\right|$ is, the edit has a greater impact on the sentence, whether beneficial or harmful.

\paragraph{Final Score}
\label{our_approach}

As shown in Figure \ref{fig1}, {given the system edit set $E$, the gold edit set $G$, and the edit score set $W$}, we treat each edit score as the corresponding edit weight and apply edit weight on each edit to compute precision, recall and F$_{0.5}$ measure respectively:

\begin{equation}
    P=\frac{\sum_{c \in E \cap G}w_c}
    {\sum_{e \in E}w_e}  
\end{equation}
\begin{equation}
    R=\frac{\sum_{c \in E \cap G}w_c}
    {\sum_{g \in G} w_g} 
\end{equation}

\begin{equation}
    F_{\beta}=(1 + {\beta}^2) \cdot \frac{P \cdot R}{({\beta}^2 \cdot P) + R} 
\end{equation}
A $\beta$ value of 0.5 is used in PT-M$^{2}$ follows vanilla M$^2$.
To compute the system score, we compute F$_{0.5}$ based on the whole corpus to represent the corpus-level PT-M$^{2}$ and use the average of F$_{0.5}$ score for each sentence to represent the sentence-level PT-M$^{2}$ \cite{napoles2016there}. Besides, since our approach is transparent to the type of edits, it can be directly applied to other overlap-based GEC metrics, such as ERRANT.

\section{Experiment}
\label{sec:experiment}
PT-M$^{2}$ can combine the advantages of both PT-based metrics and overlap-based GEC metrics. In PT-M$^{2}$, PT-based metrics only compute edit scores that can reduce the impact of unchanged part on the final score, paying more attention to the changes in the source sentences and editing scores are applied as corresponding edit weights on the M$^{2}$. 

To demonstrate the effectiveness of our approach, we use PT-M$^{2}$ to evaluate GEC system outputs and compute the correlation with human judgments on the CoNLL14 evaluation task \cite{grundkiewicz2015human}. 
We choose Pearson $\gamma$ and Spearman $\rho$ to measure the correlations.
We use PT-M$^{2}$ to compute both the system score at corpus- and sentence-level~\cite{napoles2016there}. 
We adopt two well-known unsupervised PT-based metrics BERTScore and BARTScore to score edits, which can be used in multiple domains, tasks and languages \cite{scialom2021beametrics}. The other metrics have been introduced in Section~\ref{metrics}.

\begin{table*}[t]
\centering
\begin{tabular}{crrrrrrr}
\toprule
\multirow{2}{*}{\bf Rank} &\multicolumn{4}{c}{\textbf{Metric}} & \multicolumn{2}{c}{\textbf{Human}}  \\
\cmidrule(r){2-5} \cmidrule(r){6-7} 
& \textbf{BERTScore} & \textbf{M$^{2}$} & \textbf{SentM$^{2}$} & \textbf{PT-M$^{2}$} & \textbf{EW} & \textbf{TS} \\ 
\midrule
1& INPUT ($\Uparrow$9) & CAMB ($\Uparrow$1) & CUUI ($\Uparrow$3) & \bf AMU ($\checkmark$0) & AMU & AMU\\ 
2&UFC ($\Uparrow$4) & CUUI ($\Uparrow$2) &\bf AMU ($\Downarrow$1) & CUUI ($\Uparrow$2) &RAC & CAMB\\ 
3&IITB ($\Uparrow$6) & AMU ($\Downarrow$2) &\bf CAMB ($\checkmark$0) &\bf CAMB ($\checkmark$0) &CAMB & RAC\\ 
4&\bf RAC ($\Downarrow$1) &\bf POST ($\Uparrow$1) &\bf POST ($\Uparrow$1) &\bf RAC ($\Downarrow$1) &CUUI & CUUI\\ 
5&SJTU ($\Uparrow$5) & NTHU ($\Uparrow$7) & NTHU ($\Uparrow$7) &\bf POST ($\checkmark$0) &POST & POST\\ 
6&\bf PKU ($\checkmark$0) & RAC ($\Downarrow$3) & RAC ($\Downarrow$3) &\bf PKU ($\checkmark$0) &UFC & PKU \\ 
7&AMU ($\Downarrow$6) &\bf UMC ($\checkmark$0) &\bf PKU ($\checkmark$0) & UFC ($\Downarrow$1) &PKU & UMC\\ 
8&POST ($\Downarrow$3) & PKU ($\Downarrow$1) &\bf UMC ($\checkmark$0) & SJTU ($\Uparrow$2) &UMC & UFC\\ 
9&CUUI ($\Downarrow$5) & SJTU ($\Uparrow$1) & SJTU ($\Uparrow$1) &\bf IITB ($\checkmark$0) &IITB & IITB\\ 
10&\bf UMC ($\Downarrow$2) &\bf UFC ($\Downarrow$2) &\bf UFC ($\Downarrow$2) &\bf NTHU ($\Uparrow$2) &SJTU & INPUT\\
11&IPN ($\Uparrow$2) & IPN ($\Uparrow$2) & IITB ($\Downarrow$2) & \bf INPUT ($\checkmark$0) &INPUT & SJTU\\ 
12&CAMB ($\Downarrow$9) & IITB ($\Downarrow$3) &\bf INPUT ($\Downarrow$1) & UMC ($\Downarrow$4) &NTHU & NTHU\\ 
13&NTHU ($\Downarrow$1) & INPUT ($\Downarrow$2) &\bf IPN ($\checkmark$0) &\bf IPN ($\checkmark$0) &IPN & IPN\\ \hdashline \noalign{\vskip 0.5ex}
$\Delta$ & 53 & 27 & 21 &\bf 12 &- &- \\
\bottomrule
\end{tabular}
\caption{System rankings by different metrics. $\Uparrow$/$\Downarrow$ denotes that the rank given by the evaluation metric is higher/lower than human judgments, and $\checkmark$ denotes that the given rank is equal to human ranking. The lowest rank difference in each rank is highlighted in \textbf{bold}. PT-M$^{2}$ successfully ranks the best system that the other metrics fail. Besides, it also shows the lowest rank difference ($\Delta$).}
\label{system_ranking}
\end{table*}

\subsection{Main Results}
Table~\ref{main_result} reports the correlations of overlap-based GEC metrics, PT-based metrics, and our PT-based GEC metric PT-M$^{2}$. PT-M$^{2}$ aligns better with human judgments compared to traditional overlap-based GEC metrics, either computed at corpus- or sentence-level \cite{napoles2016there}. 
Besides, for the two human rankings proposed by EW and TS, PT-M$^{2}$ makes a significant improvement on both Pearson and Spearman correlations. Although BERTScore and BARTScore are computed in different ways \cite{sai2022survey}, we find that our approach uses either of them as the edit scorer can get a similarly high correlation. 

We also test a variant of our approach, namely PT-ERRANT, with the same operations as in PT-M$^{2}$. 
Compared to ERRANT, PT-ERRANT gets higher correlation with human judgments. 
We find that comparing PT-M$^{2}$ and PT-ERRANT, PT-M$^{2}$ correlates better at the sentence-level while PT-ERRANT aligns better at the corpus-level. 
We also test different values of $\beta$ when computing F$_\beta$ as the final score. 
The results show that PT-M$^{2}$ correlates more stable and is consistently better than M$^{2}$. It is worth mentioning that the sentence-level PT-M$^{2}$ with BERTScore as the scorer gets the highest correlation among all of the metrics we have experimented with.
Therefore, we treat it as the main version of PT-M$^{2}$ and use the sentence-level PT-M$^{2}$ in the subsequent experiments.

\begin{table}[t]
\scalebox{0.93}{
\centering
{
\begin{tabular}{lcccccc}
\toprule
\multirow{2}{*}{\textbf{Models}} & \multirow{2}{*}{\textbf{Size}}& \multicolumn{2}{c}{\textbf{EW}} & \multicolumn{2}{c}{\textbf{TS}} \\ 
\cmidrule(l){3-4} \cmidrule(l){5-6} && $\gamma$ & $\rho$ & $\gamma$ & $\rho$  \\
\midrule
\multirow{3}{*}{BART} & Small & 0.883 & 0.743 & 0.869 & 0.740 \\
& Base & 0.898 & 0.813 & 0.901 & 0.841 \\
& Large & 0.899 & 0.813 & 0.906 & 0.841 \\
\midrule    
\multirow{3}{*}{BERT} & Small & 0.953 & 0.923 & 0.941 & 0.890 \\
& Base & 0.949 & 0.907 & 0.938 & 0.874\\
& Large & 0.945 & 0.918 & 0.931 & 0.879 \\
\bottomrule
\end{tabular}}}
\caption{Comparison between different PT-based models. There are no significant differences between models of different sizes.}
\label{model_size}
\end{table}

\begin{figure}[t]
\centering
\includegraphics[width=0.91\columnwidth]{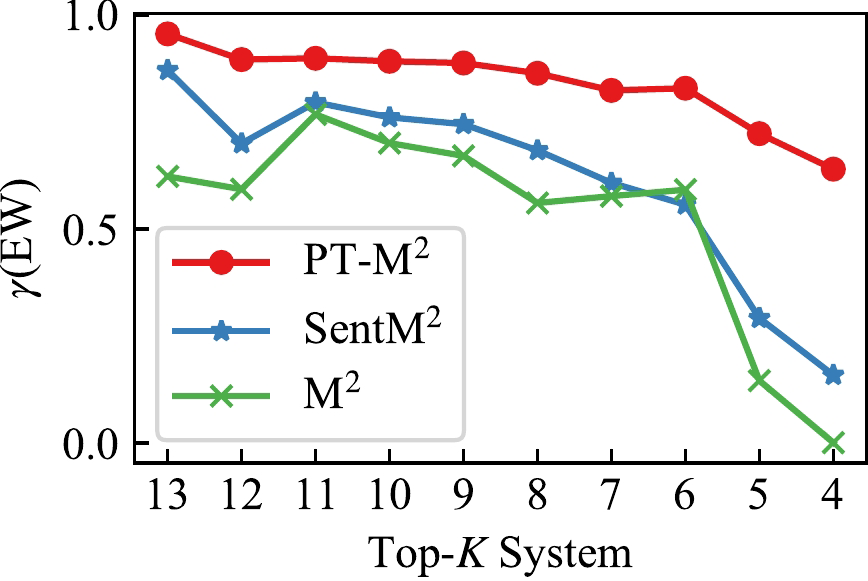}
\caption{Pearson scores of Top-$K$ system based on the EW ranking list. PT-M$^{2}$ is highly correlated with human judgments especially when all the systems are competitive (i.e., $K\leq6$).}
\label{topk_system}
\end{figure}

\subsection{Analysis}
\paragraph{Effect of PT-based models} To demonstrate the generality and generalizability of our approach, we use different sizes of PT-based metrics as the edit scorer to verify if the correlation of our approach is stable. As shown in Table \ref{model_size}, we experiment with PT-M$^{2}$ at the sentence-level, employing BERTScore and BARTScore in three different sizes, from small to large respectively. Even if we use the distilled version of BERTScore \cite{sanh2019distilbert}, PT-M$^{2}$ also aligns much better with human judgments than M$^{2}$ and gets the similar correlation with the base and even the large size of models. So we can use the smaller size of PT-based metrics to accelerate the computation of GEC evaluation, with a trivial performance drop.

\paragraph{System Ranking}
\label{ranking}
Table~\ref{system_ranking} presents the ranking results. 
Compared to other metrics, PT-M$^{2}$ shows the lowest rank differences with human judgments. 
PT-M$^{2}$ successfully ranks the best system (AMU) while the other metrics fail. 
Compared to BERTScore, PT-M$^{2}$ successfully ranks the INPUT system. 
Meanwhile, PT-M$^{2}$ gets a relatively better ranking result for the NTHU system than M$^{2}$ and SentM$^{2}$.
This confirms the effectiveness of PT-M$^2$.

\paragraph{Effect of Top-$K$ Systems} To demonstrate the effectiveness of our approach, we evaluate high-performing systems with our approach to observe the change of correlations among different metrics \cite{zhan2021difficulty}. Figure \ref{topk_system} compares the Pearson correlation of the top-$K$ systems. When $K$ decreases, the correlations of M$^{2}$ and SentM$^{2}$ are lower than before correspondingly. Meanwhile, we find that the correlation of our approach PT-M$^{2}$ computed at the sentence-level drops much slower than M$^{2}$ and SentM$^{2}$. Further, when $K$ is lower than 6, the Pearson correlations of M$^{2}$ and SentM$^{2}$ drop sharply while PT-M$^{2}$ does not act the same way under the circumstances. More specifically, as the system count drops from 13 to 4, the Pearson correlations of M$^{2}$ and SentM$^{2}$ down three times or even more. In contrast, the decline of PT-M$^{2}$ correlation is not obvious, which demonstrates the effectiveness of applying edit scores computed by PT-based metrics as corresponding edit weights on the overlap-based GEC metrics.

\begin{table*}[t]
\centering
\small
\scalebox{0.95}{
\begin{tabular}{llccc}
\toprule
\bf System &\bf Sentence & \textbf{Rank} & \textbf{M$^{2}$} & \textbf{PT-M$^{2}$} \\ 
\midrule
\textbf{SRC} & It is also \textcolor{red}{entire incorrect} to fault social media alone for \textcolor{blue}{\bf the lack} of interpersonal \textcolor{red}{skill} . 
&-       &-       \\
\textbf{REF} & It is also \textcolor{red}{\bf entirely incorrect} to fault social media alone for \textcolor{blue}{\bf the lack} of interpersonal \textcolor{red}{\bf skills} .
&- &- \\
\midrule
\textbf{NTHU} & It is also \textcolor{red}{\bf entirely incorrect} to fault social media alone for \textcolor{blue}{lack} of interpersonal \textcolor{red}{\bf skills} . & 1 
 & 0.71 & \textbf{0.88}     \\
\textbf{PKU} & It is also \textcolor{red}{entire incorrect} to fault social media alone for \textcolor{blue}{\bf the lack} of interpersonal \textcolor{red}{\bf skills} . & 2 
& \textbf{0.83} & 0.44    \\ 
\bottomrule
\end{tabular}}
\caption{Example from the CoNLL14 evaluation task. ``\textcolor{red}{\bf Red Bold}'' denotes the right correction whereas ``\textcolor{blue}{Blue Non-bolded}'' denotes the wrong one. The ranking of PT-M$^{2}$ is more in line with human judgments.}
\label{ptm2_example}
\end{table*}

\begin{figure}[t]
\centering
\includegraphics[width=0.95\columnwidth]{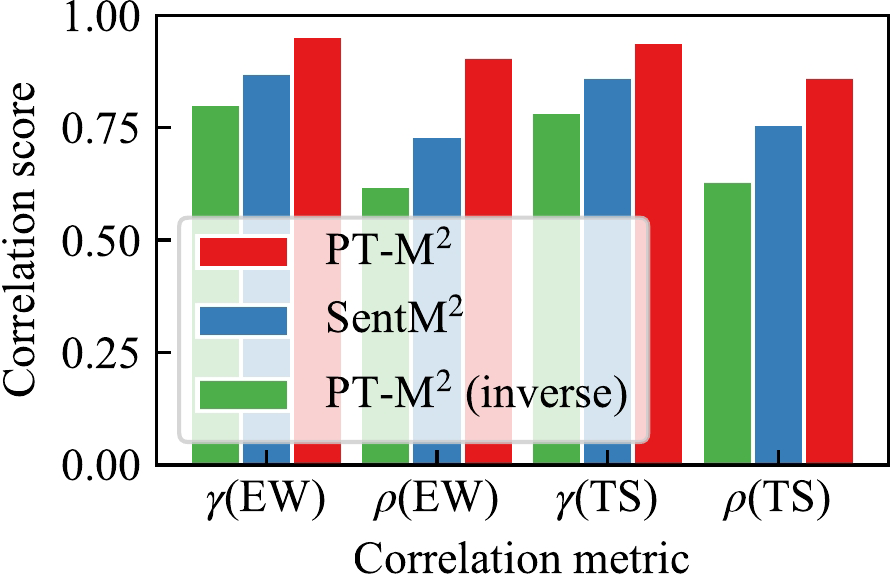}
\caption{Effectiveness of the proposed edit score. Reversing the edit score harms the correlation.}
\label{inverse_ptm2}
\end{figure}

\paragraph{Effect of Edit Score}
We carry out a set of comparative experiments to demonstrate the necessity of computing edit weights. In our approach PT-M$^{2}$, we use PT-based metrics such as BERTScore and BARTScore to compute edit scores. 
The more important the edit, the higher score we provide. 
In the comparative experiments, we use the inverse of each edit score computed by PT-based metrics as the corresponding edit weight, $ w = 1 / |\text{PTScore}(S', R) - \text{PTScore}(S, R)|$, to demonstrate the importance of edit weights on overlap-based GEC metrics. 
For convenience, we use PT-M$^{2}$ (inverse) as the above comparative approach. 

Figure \ref{inverse_ptm2} the comparison between the Pearson and Spearman correlations of the three metrics. 
After inversing edit weights computed by PT-based metrics, both the correlations measured by Pearson and Spearman are lower than SentM$^{2}$ for a large. 
Based on the above results, we demonstrate the effectiveness and advantages of employing PT-based metrics to directly compute edit weights.

\paragraph{A Case Study}
As shown in Table \ref{ptm2_example}, to explain why our approach PT-M$^{2}$ correlates better with human judgments than M$^{2}$, we present an example from the CoNLL14 evaluation task \cite{grundkiewicz2015human}. 
The human ranking shows that the NTHU system score is higher than that of the PKU system. 
M$^{2}$ provides the wrong ranking result while PT-M$^{2}$ aligns the same as human judgments. 
In this example, PT-M$^{2}$ measures the gold edit only provided by the NTHU system as the highest score, which means it is much more important to predict this edit. 
Meanwhile, PT-M$^{2}$ sets the wrong correction edit supplied by the NTHU system with a lower score, which influences the source sentence a little. 
PT-M$^{2}$ successfully scores higher for the NTHU system.

\section{Conclusion and Future Work}
\label{conclusion}
In this work, we revisit GEC metrics and explore the feasibility of employing PT-based metrics to evaluate GEC systems. 
Compared to overlap-based GEC metrics, PT-based metrics correlate worse with human judgments. 
The reason we find is that PT-based metrics compute score for each token in ungrammatical sentences, however, the scores of the unchanged part account for a large proportion of the final score, leading to an inaccurate evaluation result. 
To leverage pretrained knowledge in GEC evaluation, we propose a novel PT-based GEC metric PT-M$^{2}$, employing PT-based metrics to score edits extracted from sentence pairs and applying edit scores as corresponding edit weights on the M$^{2}$. Experiments demonstrate that PT-M$^{2}$ gets the highest correlation on the CoNLL14 evaluation task, achieving a new state-of-the-art result of 0.949 Pearson correlation. 
Besides, further analysis shows that PT-M$^{2}$ is competent to evaluate high-performing GEC systems.

In the future, we would like to test the effectiveness of PT-M$^{2}$ in the GEC evaluation tasks of other languages (e.g., Chinese). It is also worthwhile to explore the benefits of PT-M$^{2}$ as a signal for reinforcement learning to train better GEC systems.

\section*{Limitations}
Our PT-based GEC metric PT-M$^{2}$ has got the highest correlation with human judgments on GEC evaluation tasks. 
A reason that might limit the widespread use of PT-based metrics is that the calculation speed is slower than that of the traditional overlap-based GEC metrics.
As PT-M$^{2}$ uses PT-based metrics to score each edit, it takes nearly 2 minutes to calculate a system score based on a single NVIDIA GTX 3080TI card, which is slower than the calculation speed of the traditional M$^2$ score that costs nearly 10 seconds. 
Therefore, there is still room for improvement in calculation speed, and we will continue developing parallel computing to speed it up in the future.

\section*{Acknowledgments}
This work was supported in part by the National Natural Science Foundation of China (Grant No. 62206076) and Shenzhen College Stability Support Plan (Grant No. GXWD20220811173340003 and GXWD20220817123150002).
The authors would like to thank the anonymous reviewers for their insightful suggestions.

\bibliography{anthology,custom}
\bibliographystyle{acl_natbib}

\appendix
\section{Appendix}

\begin{algorithm*}[t]
\caption{Calculation Procedure of PT-M$^{2}$}
\SetKwFunction{ExtractEdit}{ExtractEdit}
\SetKwFunction{Correct}{Correct}
\SetKwFunction{ComputeScore}{ComputeScore}
\SetKwFunction{PTScore}{PTScore}
\SetKwFunction{Len}{Len}
\SetKwFunction{Max}{Max}
\SetKwProg{Fn}{Function}{:}{}
\label{algorithm1}
\KwData{source sentences $\mathbf{S}$, hypothesis sentences $\mathbf{H}$, reference sentences $\mathbf{R}$, gold edit sets $\mathbf{GOLD}$}
\KwResult{F-score F$_{0.5}$}
\Fn{\ComputeScore{$S$, $R$, $E$}}{
    initialize an empty dict $W$ ;\\
    \For{$i$ $\leftarrow$ $1$ \KwTo \Len{$E$}}
    {
        $e$ $\leftarrow$ $E_i$ ; \tcp{the $i$th edit in the edit set $E$}
        $S'$ $\leftarrow$ \Correct{$S$, $e$} ; \tcp{apply edit $e$ to correct the source sentence $S$}
        $W_e$ $\leftarrow$  |\PTScore{$S'$, $R$} - \PTScore{$S$, $R$}| ; \tcp{compute the difference of PTScore between ($S'$, $R$) and ($S$, $R$)}
        
    }
    \KwRet $W$\;
}
$F_{0.5}$ $\leftarrow$ $0$ ; \tcp{initialize $F_{0.5}$} 
\For{$i$ $\leftarrow$ $1$ \KwTo \Len{$\mathbf{S}$}}
{
    $f_{max}$ $\leftarrow$ $-1$ ; \tcp{initialize $f_{max}$}
    $S$, $H$, $R$, $GOLD$, $\leftarrow$ $\mathbf{S}_{i}$, $\mathbf{H}_{i}$, $\mathbf{R}_{i}$, $\mathbf{GOLD}_{i}$; \\
    \For{$j \leftarrow 1$ \KwTo \Len{$R$}}
    {
        $G$ $\leftarrow$ $GOLD_j$ ; \tcp{the jth gold edit set of $S$}
        $E$ $\leftarrow$ \ExtractEdit{$S$, $H$, $G$}; \tcp{extract system edit}
        $C$ $\leftarrow$ $E \cap G$ ; \tcp{choose the correct edit that system provides}
        $W$ $\leftarrow$ \ComputeScore{$S$, $R$, $E \cup G$}; \tcp{compute the score of each
        edit}
        $p$ $\leftarrow$ $\sum_{c \in C}W_c / \sum_{e \in E} W_e$ ; \tcp{compute the precision score}
        $r$ $\leftarrow$ $\sum_{c \in C}W_c / \sum_{g \in G} W_g$ ; \tcp{compute the recall score}
        $f$ $\leftarrow$ $(1 + 0.5^2) \cdot p \cdot r / (0.5^2 \cdot p + r)$ ; \tcp{compute the f score}
        $f_{max}$ $\leftarrow$ \Max{$f_{max}$, $f$}
    }
    $F_{0.5}$ $\leftarrow$ $F_{0.5} + f_{max}$
}
$F_{0.5}$ $\leftarrow$ $F_{0.5}$ $/$ \Len{$\mathbf{S}$}
\end{algorithm*}

\subsection{Algorithm}
\label{algorithm}
Algorithm \ref{algorithm1} illustrates the calculation procedure of our PT-based GEC metric PT-M$^{2}$, showing the whole process of how to compute the sentence-level PT-M$^{2}$. 

\subsection{Command-line Interface}
\label{cli}
We introduce how to compute PT-M$^{2}$ and its variant PT-ERRANT in different settings with our code:
\begin{python}
python evaluate.py
    --base [m2|sentm2|errant|senterrant] 
    --scorer [self|bertscore|bartscore]
    --source <src_file>
    --hypothesis <hyp_file>
    --reference <ref_file>
    --output <out_file>
\end{python}

\end{document}